\documentclass[letterpaper, 10 pt, conference]{ieeeconf} 

\IEEEoverridecommandlockouts 

\usepackage{graphics} 
\usepackage{epsfig} 
\usepackage{mathptmx} 
\usepackage{amsmath} 
\usepackage{amssymb} 
\usepackage{booktabs} 
\usepackage{subcaption}
\usepackage[hyphens]{url}
\usepackage{multirow}
\usepackage{balance}

\title{\LARGE \bf Benchmarking Multi-View BEV Object Detection with Mixed Pinhole and
Fisheye Cameras}

\author{Xiangzhong Liu$^{1}$ and Hao Shen$^{2}$
\thanks{$^{1}$Xiangzhong Liu and $^{2}$Hao Shen are with Machine Learning Group,
fortiss GmbH, Guerickestraße 25, 80805 Munich, Germany
{\tt\small xiangzhong.liu@tum.de}, {\tt\small shen@fortiss.org}}%
}
\usepackage{textcomp}
\usepackage[absolute,showboxes]{textpos}
\TPMargin{4pt}
\newcommand{\copyrightstatement}{
    \begin{textblock}{0.84}(0.08,0.938)
         \noindent{\footnotesize{\copyright 2026 IEEE.
         Personal use of this material is permitted.
         Permission from IEEE must be obtained for all other uses, in any current or future media, including reprinting/republishing this material for advertising or promotional purposes, creating new collective works, for resale or redistribution to servers or lists, or reuse of any copyrighted component of this work in other works.

         \noindent
         Accepted for publication in Proceedings of the IEEE International Conference on Robotics and Automation (ICRA),  Vienna, Austria, 1-5 June 2026.}}
    \end{textblock}
}  
\begin{document}
    \maketitle
    \copyrightstatement
    \thispagestyle{empty}
    \pagestyle{empty}

    \begin{abstract}
        Modern autonomous driving systems increasingly rely on mixed camera configurations
        with pinhole and fisheye cameras for full view perception. However, Bird's-Eye View
        (BEV) 3D object detection models are predominantly designed for pinhole cameras,
        leading to performance degradation under fisheye distortion. To bridge this
        gap, we introduce a multi‑view BEV detection benchmark with mixed
        cameras by converting KITTI‑360 into nuScenes format. Our study encompasses
        three adaptations: rectification for zero-shot evaluation and fine-tuning
        of nuScenes-trained models, distortion-aware view transformation modules
        (VTMs) via the MEI camera model, and polar coordinate representations to
        better align with radial distortion. We systematically evaluate three representative
        BEV architectures, BEVFormer, BEVDet and PETR, across these strategies. We
        demonstrate that projection-free architectures are inherently more robust
        and effective against fisheye distortion than other VTMs. This work
        establishes the first real-data 3D detection benchmark with fisheye and
        pinhole images and provides systematic adaptation and practical
        guidelines for designing robust and cost-effective 3D
        perception systems. The code is available at \url{https://github.com/CesarLiu/FishBEVOD.git}.
    \end{abstract}

    \section{INTRODUCTION}
    \label{sec:intro}

    Current BEV multi-view 3D object detection (3DOD) systems have achieved
    remarkable success on standardized datasets like nuScenes~\cite{caesar2020nuscenes},
    where uniform pinhole camera configurations provide consistent geometric properties
    for all views. However, this setup diverges from real-world autonomous driving
    systems that adopt mixed camera configurations for cost-effectiveness.
    Serial production vehicles combine pinhole cameras for forward-facing perception
    with fisheye cameras for surround-view visualization,
    near-field parking assistance, and 2D object recognition rather than 3D perception
    tasks~\cite{kumar2023surround}, achieving $360^{\circ}$ coverage at a
    fraction of the cost of an all-pinhole system.

    \begin{figure}[htbp]
        \centering
        \includegraphics[width=1.0\linewidth]{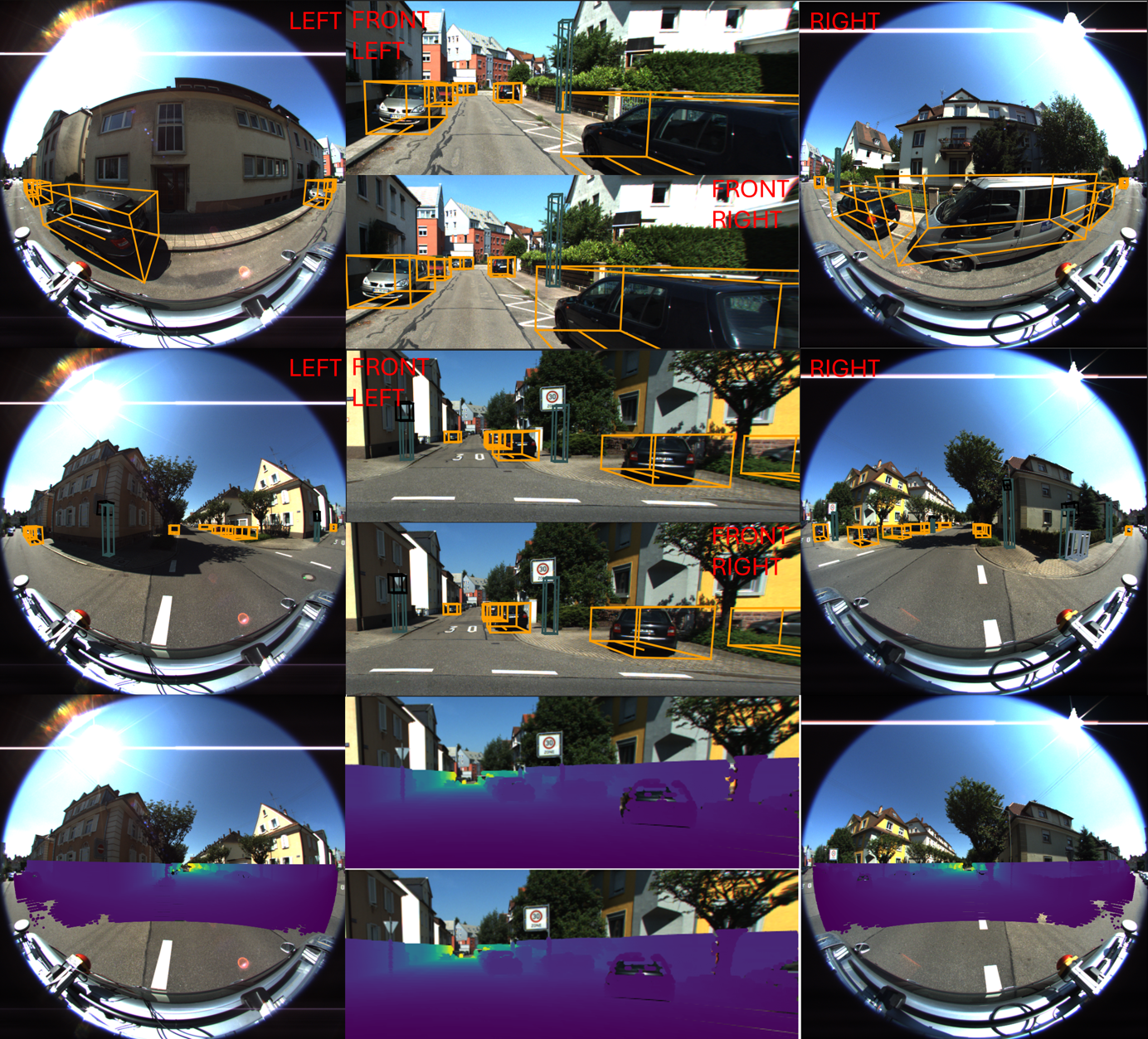}
        \caption{Qualitative detection results on KITTI-360 mixed camera
        configuration. 3D bounding box and point cloud rendering overlaid on pinhole
        and fisheye images. The visualization is created with a customized
        nuScenes devkit. Our method successfully handles the severe radial
    distortion in fisheye cameras while maintaining consistent detection
    accuracy across different camera types.}
        \label{fig:qualitive_viz_comp}
    \end{figure}

    While BEV methods excel on pinhole cameras~\cite{li2022bevsurvey}, their
    performance remains largely underexplored on mixed configurations, especially
    regarding fisheye cameras. Fisheye images present several fundamental
    challenges:
    \begin{enumerate}
        \item Severe radial distortion causes non-uniform spatial sampling and non-linear
            transformations, leading to inconsistent object scales and feature
            representations.

        \item The wide field of view compresses distant objects, introducing
            ambiguity and information loss in BEV projections.

        \item Depth estimation is unreliable in highly distorted regions,
            hindering accurate 3D lifting.
    \end{enumerate}

    These issues fundamentally disrupt the spatial consistency assumptions
    underlying BEV models, leading to significant performance degradation of existing
    BEV architectures in 3D perception tasks~\cite{samani2023f2bev}. Due to the
    lack of large-scale datasets with fisheye imagery and reliable 3D
    annotations, this issue has remained largely unaddressed in current literature.

    The KITTI-360 dataset~\cite{liao2022kitti} presents a unique opportunity to address
    this gap, with a comprehensive sensor suite including two $180^{\circ}$
    fisheye cameras alongside forward-facing stereo cameras, creating a simplified
    mixed configuration that resembles real-world deployments. While KITTI-360
    has been utilized for 3D semantic segmentation and monocular object
    detection, we identify our work as the first to systematically evaluate multi-view
    BEV-based 3DOD on a real-world dataset with mixed imagery.

    While rectification offers conceptual simplicity, it introduces
    computational overhead and information loss through cropping. Direct fisheye
    modeling preserves the full FoV advantages while eliminating rectification overhead,
    allowing models to adapt to non-linear projections by learning~\cite{samani2023f2bev}.
    We investigate three representative BEV architectures: PETR~\cite{liu2022petr}
    (projection-free), BEVFormer~\cite{li2024bevformer} (backward projection),
    and BEVDet~\cite{huang2021bevdet} (forward projection). We propose polar coordinate
    transformation to align the BEV representation with the fisheye distortion,
    as polar grids naturally capture the angular consistency and radial
    compression inherent to wide-angle lenses~\cite{jiang2023polarformer,yu2024polarbevdet}.
    Our contributions are summarized as follows:
    \begin{itemize}
        \item \textbf{First Multi-View Benchmark on KITTI-360:} We establish
            KITTI-360 as a comprehensive benchmark for mixed camera BEV object
            detection, developing conversion pipelines to enable standardized evaluation
            with the nuScenes devkit.

        \item \textbf{Systematic Architecture Adaptation:} We implement both
            geometric rectification and distortion modeling with MEI camera
            model across three VTMs, providing the first systematic comparison of
            these paradigms.

        \item \textbf{Polar Representation tailored for Fisheye Geometry:} We
            introduce polar coordinate transformations for fisheye camera MEI
            model, better reflecting the fisheye radial structure and enabling geometry-aware
            feature alignment in the BEV space.
    \end{itemize}

    While recent advances such as temporal modeling~\cite{huang2022bevdet4d,wang2023exploring,yang2023bevformer}
    and 2D auxiliary detection heads~\cite{wang2023focal} have proven effective for
    general BEV pipelines, our study focuses specifically on the geometric
    challenges posed by distortion.

    \section{RELATED WORK}

    \label{sec:related_works}

    \subsection{BEV Multi-View 3D Detection}

    BEV representation has emerged as the dominant paradigm for 3DOD in autonomous
    driving~\cite{li2022bevsurvey}, transforming multiple perspective views into
    a unified top-down coordinate system. Based on their view transformation modules,
    current approaches are categorized into three families~\cite{li2023fb}:

    \textbf{Forward Projection.} LSS~\cite{philion2020lift} pioneered this paradigm
    by predicting per-pixel depth distributions, then lifting 2D features into
    3D space via geometric projection. BEVDet~\cite{huang2021bevdet} and BEVFusion~\cite{liu2022bevfusion}
    extend this approach but remain vulnerable to depth estimation errors—particularly
    problematic for geometrically distorted fisheye imagery.

    \textbf{Backward Projection.} BEVFormer~\cite{li2024bevformer} predefines 3D
    coordinates and projects reference points back to 2D images through
    deformable cross-attention, enabling denser BEV representations. BEVFormerv2~\cite{yang2023bevformer}
    improved efficiency through perspective supervision and temporal modeling.

    \textbf{Projection-Free.} PETR~\cite{liu2022petr} bypasses explicit geometric
    projection by enriching 2D features with 3D positional embeddings derived from
    camera parameters. StreamPETR~\cite{wang2023exploring} extended this with
    temporal cues. While computationally efficient, these methods rely on
    learned geometric priors potentially sensitive to camera configuration
    changes.

    \textbf{Polar BEV Representation.} Traditional multi-view 3DOD methods employ
    Cartesian BEV representations with uniform grid rasterization, creating a
    structural mismatch with the natural radial distribution of camera frustum
    information. PolarFormer~\cite{jiang2023polarformer} and PolarBEVDet~\cite{yu2024polarbevdet}
    replace rectangular representations with polar coordinates $(r, \theta)$ for
    more coherent multi-view feature aggregation.

    These methods excel on pinhole-camera datasets like nuScenes but inherently assume
    linear projection models, making their performance on real-world fisheye
    images with severe distortions largely unexplored and potentially
    problematic.

    \subsection{Fisheye Camera Modeling in 3D Perception}

    The integration of fisheye cameras into 3D perception pipelines presents critical
    challenges due to severe radial distortion that violates the linear perspective
    assumptions underlying standard computer vision algorithms~\cite{kumar2023surround}.
    Fisheye lenses achieve ultra-wide fields of view (typically $180^{\circ}$ or
    greater) through intentional geometric distortion~\cite{kannala2006generic},
    creating non-uniform spatial sampling, which violates the translational
    equivariance of standard CNNs.

    Traditional approaches address fisheye distortion via rectification, mapping
    distorted images to approximate pinhole views~\cite{kumar2023surround}. This
    preprocessing enables direct application of existing BEV models without
    architectural modifications, but introduces computational overhead,
    inevitable information loss through cropping and non-uniform resolution
    distribution that affect feature quality for downstream tasks. Recent methods
    bypass rectification by directly integrating fisheye camera models in learning-based
    approaches. F2BEV~\cite{samani2023f2bev} pioneered this approach by
    replacing BEVFormer's pinhole projection with the MEI camera model~\cite{mei2007single}.
    Beyond that, spherical transformer architectures~\cite{carlsson2024heal}
    extend Vision Transformers to spherical projections, while DarSwin~\cite{athwale2023darswin}
    presents a distortion-aware encoder-only architecture, offering generalized
    distortion modeling. RectConv~\cite{griffiths2024adapting} introduces distortion-aware
    rectified convolution that can handle camera distortion effects directly within
    the convolution operation. These models eliminate rectification loss while
    preserving the full field of view, but require geometry-aware coordinate and
    sampling adaptations.

    \subsection{Benchmarks with Fisheye Imagery}

    Current real-world datasets primarily target 2D perception tasks, reflecting
    the historical limitations of fisheye cameras for situational awareness.
    WoodScape~\cite{yogamani2019woodscape} provides 40K fisheye images but lacks
    public 3D annotations. SynWoodScape~\cite{sekkat2022synwoodscape} addresses 3D
    annotation absence through synthetic data generation, though its real-world
    transferability remains questionable, with only 500 samples publicly available. FB-SSEM~\cite{samani2023f2bev}
    provides a Unity-generated synthetic dataset featuring 20,000 samples of fisheye
    SVC images with corresponding BEV maps for simulated parking lot scenarios.
    Major autonomous driving benchmarks (nuScenes~\cite{caesar2020nuscenes},
    Waymo~\cite{sun2020scalability}) exclusively use pinhole cameras, leaving
    mixed-camera evaluation underexplored.

    KITTI-360~\cite{liao2022kitti} uniquely provides both forward-facing stereo
    pinhole cameras and dual $180^{\circ}$ fisheye cameras with high-quality
    LiDAR-derived 3D annotations. However, it has been primarily used for 3D
    semantic segmentation with limited BEV detection exploration due to the format
    incompatibilities with nuScenes framework.

    This work addresses this by developing the first systematic KITTI-360 to
    nuScenes conversion pipeline, establishing a standardized mixed-camera benchmark
    for BEV 3DOD. Our contribution enables the direct application of nuScenes-compatible
    models to fisheye images, fostering research into practical perception systems
    with mixed or pure fisheye cameras.

    \section{METHODOLOGY}

    \label{sec:methodology}

    Our core methodological contribution is a framework for the adaptation of existing
    BEV-based 3DOD models to mixed pinhole and fisheye camera configurations and evaluation
    on a real-world dataset. Our approach is three-pronged. First, to enable
    rigorous and standardized evaluation, we establish a new multi-view benchmark
    by converting the KITTI-360 dataset into the nuScenes format. Second, we perform
    a systematic adaptation of three representative VTMs with MEI for unified camera
    modeling. Additionally, we introduce polar coordinate transformations that naturally
    align with fisheye geometry to improve spatial feature aggregation.

    \subsection{Dataset Conversion and Rectification}

    While KITTI-360 provides rich urban scene annotations with comprehensive
    $360^{\circ}$ sensor coverage for both cameras and LiDARs~\cite{liao2022kitti},
    existing BEV detection methods are predominantly designed around nuScenes data
    structure. The format conversion enables the first multi-view evaluation on
    European urban driving scenarios, complementing existing nuScenes evaluations
    from America and Asia.

    \paragraph{Conversion Pipeline}
    The XML/pose annotations differ significantly from the tokenized JSON
    structure of nuScenes and static objects are only labeled once globally
    without frame assignments. We convert KITTI-360 to nuScenes format in three key
    steps:
    \begin{enumerate}
        \item \textbf{Scene partitioning and split definitions} - We adopt the
            official train/val splits and partition long sequences into scene
            windows (200m) aligned with nuScenes.

        \item \textbf{Sample identification} - Keyframes are defined by matching
            accurate georegistered vehicle poses to synchronized data frames, excluding
            non-keyframes to maintain dense annotations.

        \item \textbf{Annotation conversion} - Static objects are assigned to
            frames via distance and LiDAR-visibility checks, and objects are transformed
            into nuScenes’ center–size–quaternion format. CityScapes' label definitions
            are mapped to nuScenes’ 10 detection classes, with extra KITTI-360
            categories preserved as extensions (e.g. pole and traffic signs).
    \end{enumerate}
    
    \paragraph{Geometric Rectification}

\begin{figure}[htbp]
        \centering
        \includegraphics[width=1.0\linewidth]{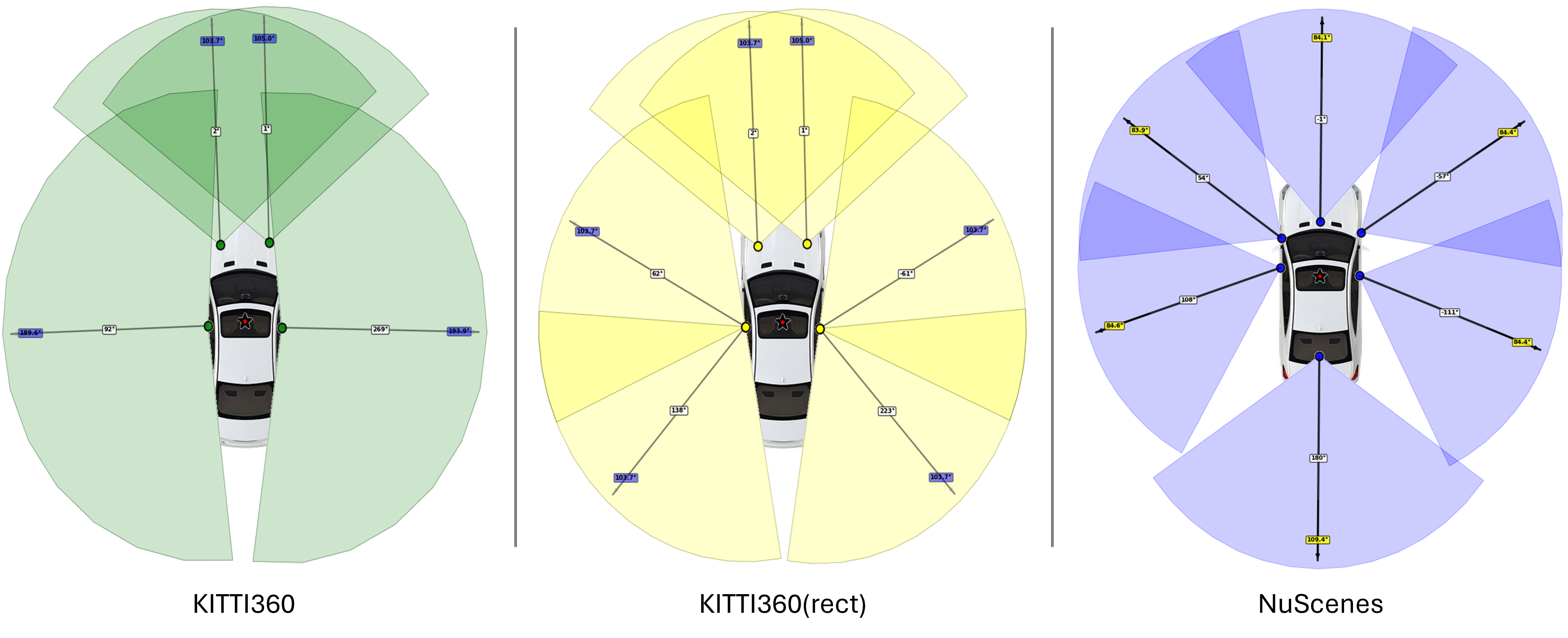}
        \caption{Camera configuration and full field-of-view coverage comparison.
        Left: Original KITTI-360 with stereo cameras and fisheye cameras. Middle:
        Rectified KITTI-360 with 6 pinhole cameras. Right: NuScenes
        configuration with 6 pinhole cameras.}
        \label{fig:fov_kitti}
    \end{figure}

    For comparison, we establish a rectification baseline by transforming fisheye
    cameras into virtual pinhole views. Each fisheye generates two
    cameras with the same focal length as the front ones: forward ($30^{\circ}$ rotation) and backward-facing ($-46^{\circ}$
    rotation), with $-4^{\circ}$ downward pitch. Together with the stereo pinhole
    cameras, this creates a 6-camera setup comparable to nuScenes as shown in
    Figure~\ref{fig:fov_kitti}. We evaluate this rectified dataset in two ways: zero-shot
    inference with nuScenes-trained models to measure transferability, and fine-tuning
    to identify the domain gap and establish performance upper
    bounds. While rectification enables compatibility with existing BEV detectors,
    it also suffers from reduced FoV, resolution distortion and computational overhead,
    motivating the direct fisheye-aware approaches.

    \subsection{Distortion-Aware View Transformation}
    We adopt the unified MEI camera model~\cite{mei2007single} to handle mixed pinhole
    and fisheye lens distortion in the KITTI-360 dataset. Unlike traditional
    pinhole cameras that assume rectilinear projection, fisheye cameras exhibit significant
    radial distortion requiring specialized modeling. The MEI model provides a
    mathematically elegant framework unifying pinhole, fisheye, and catadioptric
    cameras through a single parameter set, enabling seamless integration across
    different camera types within the same multi-view system~\cite{samani2023f2bev}.

    For a 3D point $(X, Y, Z)$ in camera coordinates, the MEI model first
    projects the point to a unit sphere, then applies perspective projection with
    mirror parameter $\xi$, followed by radial distortion correction and image
    plane projection with $\mathbf{K}_{f}$.
    \begin{equation}
        \begin{aligned}
            \mathbf{P}_{s} & = \mathbf{P}/||\mathbf{P}||                                         \\
            \mathbf{P}_{c} & = \left(\frac{X_{s}}{Z_{s} + \xi}, \frac{Y_{s}}{Z_{s} + \xi}\right) \\
            r^{2}          & = X_{c}^{2}+ Y_{c}^{2}                                              \\
            \mathbf{P}_{d} & = (1 + k_{1}r^{2}+ k_{2}r^{4}) \times \mathbf{P}_{c}                \\
            \mathbf{P}_{I} & = \mathbf{K}_{f}\mathbf{P}_{d}
        \end{aligned}
    \end{equation}
    This unified formulation reduces to pinhole projection when $\xi = 0$ and $k_{i}
    = 0$, enabling consistent processing across mixed camera configurations.

    \begin{figure*}[htbp]
        \centering
        \includegraphics[width=1.0\linewidth]{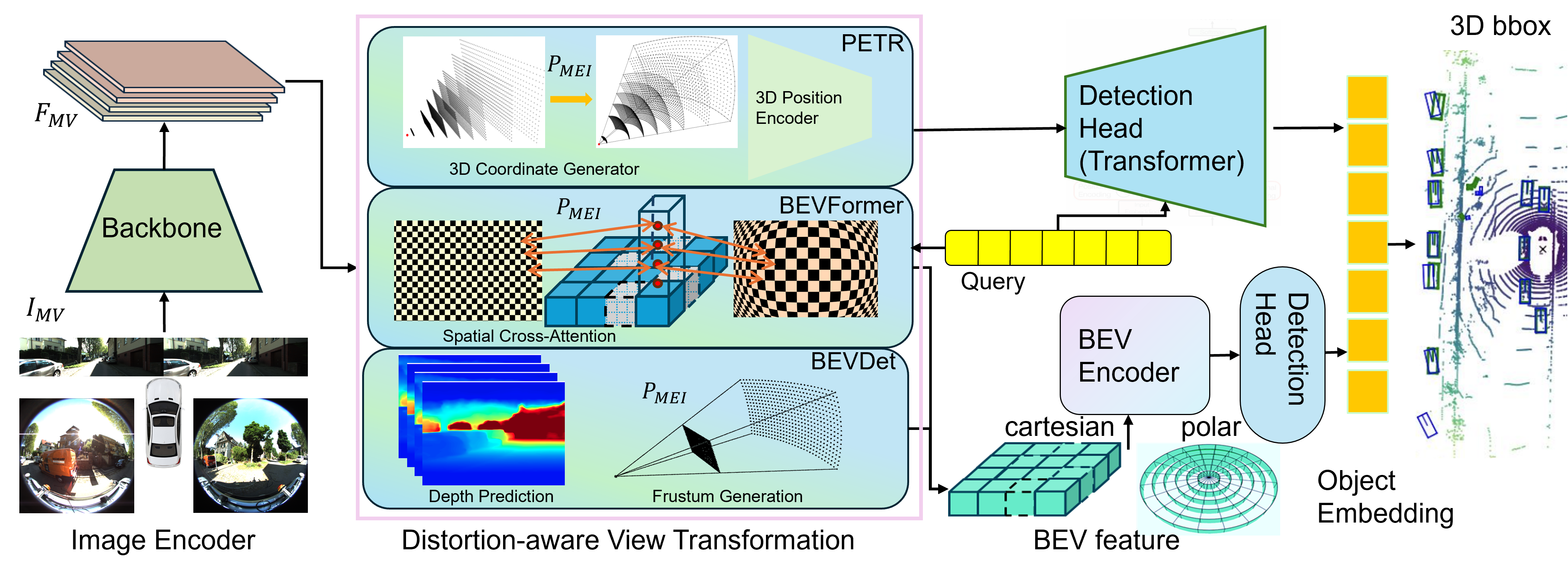}
        \caption{Overview of distortion-aware BEV 3D object detection
        framework. Multi-view images (pinhole+fisheye) are processed by a shared
        backbone encoder, then fed into three distortion-aware view
        transformation modules via MEI camera model. The resulting BEV features can
        be represented in either Cartesian or polar coordinates to better align with
        fisheye geometry. A detection head(Transformer or CNN) processes BEV features to produce final 3D object detection outputs.}
        \label{fig:full_structure}
    \end{figure*}
    \paragraph{3D Position Encoding for PETR}

    PETR employs 3D position encoding to establish spatial correspondences
    between image features and 3D queries. We replace the reference points generation
    with our distortion-aware ray generation process in the position encoding module
    to account for non-linear distortion characteristics:
    \begin{equation}
        \mathbf{R}_{\text{f}}(u,v,d) = \text{UnprojectMEI}([u,v], \mathbf{K}_{\text{f}}
        , \xi, k_{1}, k_{2}) \times \mathbf{D}_{\text{u}}
    \end{equation}

    where $\text{UnprojectMEI}(\cdot)$ implements the inverse unified camera model
    transformation along uniformly distributed depth $\mathbf{D}_{\text{u}}$.
    The 3D position encoding queries $\mathbf{Q}_{\text{pos}}$ are subsequently
    computed as:

    \begin{equation}
        \mathbf{Q}_{\text{pos}}= \text{PE}_{\text{fpe}}(\mathbf{R}_{\text{f}}(u,v
        ,d))
    \end{equation}

    where $\text{PE}_{\text{fpe}}$ denotes the feature-guided positional encoding
    (FPE) function~\cite{liu2023petrv2}. The FPE mechanism makes PE data-dependent
    by leveraging features learned from raw images to adaptively modulate the
    positional embedding. Specifically, a small MLP $\phi$ processes the
    projected 2D features from the distortion-aware ray
    $\mathbf{R}_{\text{f}}(u,v,d)$, yielding attention weights that adapt according
    to the underlying distorted image content. This feature-dependent encoding
    helps the model implicitly consider depth cues and geometric relationships
    that are particularly challenging in distorted image regions. By incorporating
    fisheye-specific geometric priors through the MEI unprojection, the
    downstream cross-attention layers attend along geometrically accurate
    fisheye rays rather than erroneous straight rays ignoring radial distortion,
    while the FPE mechanism ensures that positional embeddings correctly capture
    the non-uniform spatial sampling and varying information density characteristic
    of distorted imagery.

    \paragraph{Spatial Cross-Attention in BEVFormer}

    BEVFormer builds BEV representations by applying spatial cross-attention between
    3D BEV queries and 2D image features. We introduce a distortion-aware
    spatial cross-attention module that explicitly models non-linear projection
    effects of distortion in the reference point sampling (RPS) module for better
    spatial alignment.

    The RPS is reformulated as:
    \begin{equation}
        \mathbf{p}_{\text{ref}}= \text{ProjectMEI}(\mathbf{q}_{\text{bev}}+ \Delta
        \mathbf{p}, \mathbf{K}_{\text{fisheye}}, \xi, k_{1}, k_{2})
    \end{equation}

    where BEV queries $\mathbf{q}_{\text{bev}}$ are first lifted to 3D space and
    $\text{ProjectMEI}(\cdot)$ applies the unified camera model to project 3D
    reference points to 2D image feature space as a primary alignment. The sampling
    offset $\Delta\mathbf{p}$ is predicted by the deformable attention mechanism
    accounting for the non-uniform information density in fisheye images,
    particularly in edge regions where distortion is strong:
    \begin{equation}
        \text{Attn}(\mathbf{q}_{\text{bev}}, \mathbf{f}_{\text{img}}) = \sum_{k=1}
        ^{K}W_{k}\cdot \mathbf{f}_{\text{img}}(\mathbf{p}_{\text{ref}_k}) \cdot \gamma
        (\mathbf{p}_{\text{ref}_k})
    \end{equation}

    where $W_{k}$ are the learned attention weights, $K$ is the number of sampling
    points, and $\gamma(\mathbf{p}_{\text{ref}_k})$ represents distortion-aware
    geometric weighting.

    \paragraph{Depth-Based Lifting of BEVDet}

    Unlike PETR's projection-free approach that relies on learned positional
    embeddings, BEVDet performs view transformation through explicit depth estimation.
    This architectural difference necessitates MEI integration directly into the
    geometric transformation pipeline rather than the positional encoding stage.

    We modify BEVDet's VTM to handle fisheye distortion by incorporating the
    MEI camera model into the depth-based lifting operation as PETR to
    generate 3D frustums $\mathbf{P}_{3D}$, but with $\mathbf{D}_{\text{pred}}$,
    which is predicted with the depth estimation head along 3D rays that account for
    radial distortion. The subsequent BEV projection maintains the original formulation:
    \begin{equation}
        \mathbf{P}_{\text{BEV}}= \text{ProjectBEV}(\mathbf{P}_{3D}, \mathbf{T}_{\text{cam2ego}}
        )
    \end{equation}

    This modification preserves BEVDet's explicit depth supervision and LSS design
    philosophy while ensuring geometrically accurate feature lifting under distortion.
    By addressing the fundamental challenge of unreliable depth cues in heavily
    distorted regions through proper geometric modeling, this approach enables BEVDet
    to maintain its depth-based advantages.

    \subsection{Polar Coordinate Transformation}

    Recognizing that the uniform-sampling assumption in Cartesian
    representations is invalid for fisheye imagery, we redesign the BEV parameterization
    in cylindrical coordinates $(\rho ,\theta,z)$. In an equidistant fisheye
    model~\cite{kannala2006generic}, pixels sample equal azimuth increments, yet
    those increments cover large ground‑plane spans near the image center and
    small spans near the rim. A polar grid representation aligns with fisheye geometry
    by:
    \begin{itemize}
        \item Preserving angular consistency across camera frustums,

        \item Enforcing uniform sampling density in radial and angular dimensions,
            and

        \item Enabling more natural feature aggregation for wide-angle perception
    \end{itemize}

    \paragraph{Polar-PETR}

    We re-parameterize PETR's 3D position encoding and object queries from Cartesian
    $(x,y,z)$ to cylindrical coordinates to better align with fisheye geometry.
    For each 3D point generated by PETR's frustum sampling, we compute its polar
    position $(\rho,\theta,z)$ and normalize with the maximum detection range $\rho
    _{\max}$ and angular range $( 2\pi)$, which are further fed into the 3D
    position encoder. To handle the angular wrap-around at $\theta = 0/ 2\pi$,
    we deploy sinusoidal positional encoding, $[\sin \theta, \cos\theta, \rho_{\text{norm}}
    , z_{\text{norm}}]$ ensuring continuous embeddings across the angular boundary.

    Object queries (anchors) are initialized uniformly in polar space and directly
    fed to the transformer decoder without conversion, as we expect the
    transformer to learn polar representations and predict spatial offsets to
    the polar reference points. This approach places queries on radial beams at
    fixed angular increments, naturally exploiting multi-view symmetry where 
    objects at the same radius appear similarly across different camera views. However,
    for loss calculation and final box regression in the detection head, predicted
    $(\rho,\theta,z)$ coordinates and offsets are converted back to Cartesian coordinates,
    to ensure compatibility with standard 3D detection evaluation protocols and
    ground truth annotations.

    \paragraph{Polar-BEVFormer}

    Within BEVFormer’s spatial cross‐attention, reference points sampled from the
    BEV grid are likewise re‐parameterized from Cartesian to polar coordinates.

    The reference point generator creates normalized polar coordinates that serve
    as 3D anchors for deformable attention sampling. For camera projection, polar
    coordinates are converted to Cartesian form before applying standard camera intrinsics
    and extrinsics. This conversion occurs within the spatial cross-attention
    module in the Transformer encoder, allowing existing projection code to remain unchanged
    while enabling polar BEV queries to attend to appropriate image regions.

    BEV query embeddings are enhanced with polar-aware positional encoding.
    Instead of Cartesian coordinates, we encode each query's $(\rho,\theta)$ position
    using sinusoidal functions applied to normalized polar coordinates, ensuring
    that the transformer captures radial symmetry and angular relationships
    inherent to fisheye geometry while maintaining spatial consistency across
    different radial distances and angular orientations.

    \paragraph{PolarBEVDet}
    We build upon the existing PolarBEVDet implementation \cite{yu2024polarbevdet},
    which adapts the Lift‑Splat‑Shoot pipeline into a polar grid. Frustum points
    $(u, v, d)$ are back‐projected into polar coordinates with modeling the distortion
    via MEI model, then “splatted” into $N_{\theta}\! \times\! N_{\rho}$
    angular‐radial bins via simple indexing and sum‐pooling. Features are aggregated
    via sum pooling into a regular feature map $F \in \mathbb{R}^{C \times N_{\theta}
    \times N_{r}}$, naturally aligning with fisheye geometry while preserving
    full field-of-view coverage. This plug‐and‐play implementation provides a polar
    BEV map with minimal changes to the original depth‐lifting implementation.

    \begin{figure}[htbp]
        \centering
        \includegraphics[width=1.0\linewidth]{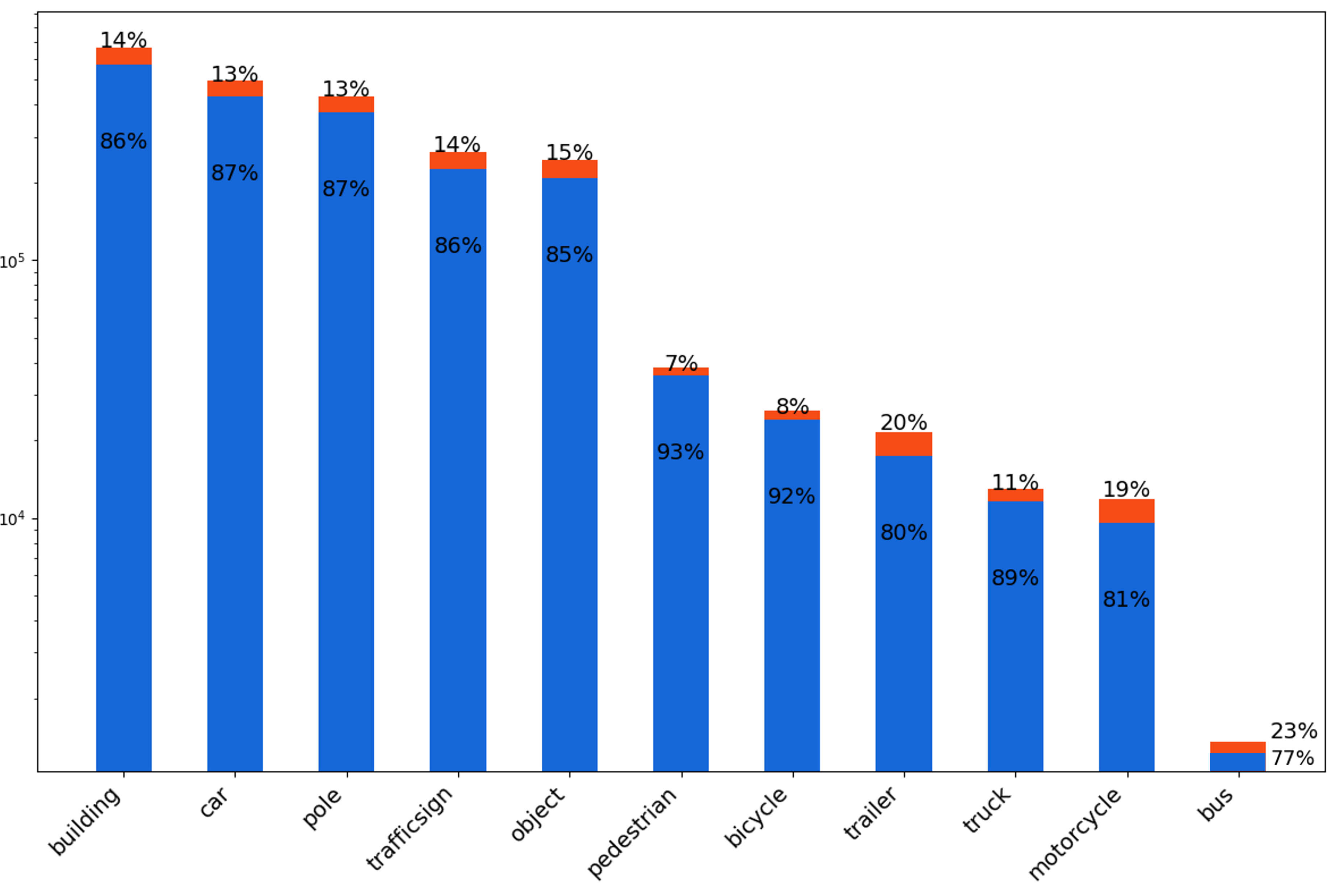}
        \caption{Class distribution in KITTI-360 dataset (log scale). The annotation
        distribution exhibits significant class imbalance, heavily skewed toward
        static infrastructure objects, while dynamic objects relevant to autonomous
        driving other than car represent a smaller portion. Blue bars indicate
        training samples, orange bars for validation, with consistent ratios
        maintained across splits.}
        \label{fig:distribution_stat}
    \end{figure}

    \section{EXPERIMENTS}

    \label{sec:experiments} To validate our proposed distortion-aware adaptation
    framework, we conduct a series of experiments designed to answer three key
    questions:
    \begin{itemize}
        \item What is the performance degradation when SotA BEV models are naively
            applied to fisheye data?

        \item How effectively does our direct adaptation approach, using the MEI
            camera model, improve performance for different model paradigms?

        \item Does our proposed polar coordinate positional encoding provide additional
            benefits for the most suitable architecture?
    \end{itemize}

    \subsection{Experimental Setup}

    \paragraph{Dataset and Preprocessing}
    The converted KITTI-360 dataset comprises 55,526 training samples from 258
    distinct scenes, and we validate on 8,554 samples from 41 scenes that are geographically
    separate from the training scenes to ensure fair evaluation of generalization.
    The dataset exhibits significant class imbalance, as shown in Figure~\ref{fig:distribution_stat}.
    While the training and validation splits maintain consistent class ratios, the
    annotation distribution is heavily skewed toward static objects
    (buildings, poles and traffic signs) compared to movable objects for
    autonomous driving.
    \begin{table*}
        [!htbp]
        \centering
        \caption{Comprehensive evaluation on KITTI-360 fisheye benchmark. Results
        show zero-shot performance of nuScenes-trained models, baseline training
        ignoring distortion, our distortion-aware adaptations with MEI camera model,
        polar coordinate enhancements, and upper-bound performance on rectified data.}
        \label{tab:comprehensive_results}
        \begin{tabular}{l|c|c|cccccc|ccc}
            \toprule[2pt] \textbf{Model} & \textbf{Backbone} & \textbf{Method} & \textbf{mAP}$\uparrow$ & \textbf{NDS}$\uparrow$ & \textbf{mATE}$\downarrow$ & \textbf{mASE}$\downarrow$ & \textbf{mAOE}$\downarrow$ & \textbf{mAVE}$\downarrow$ & \textbf{AP}$_{\text{car}}$ $\uparrow$ & \textbf{AP}$_{\text{ped}}$ $\uparrow$ & \textbf{AP}$_{\text{bus}}$ $\uparrow$ \\
            \midrule                     &                   & Zero-Shot       & 0.008                  & 0.030                  & 1.036                     & 0.830                     & 1.139                     & 0.957                     & 0.055                                 & 0.000                                 & 0.000                                 \\
                                         &                   & Baseline        & 0.121                  & 0.159                  & 0.736                     & 0.481                     & 1.031                     & 1.017                     & 0.462                                 & 0.096                                 & 0.000                                 \\
            BEVDet                       & ResNet-50         & DA-VTM          & 0.150                  & 0.212                  & 0.686                     & 0.400                     & 1.089                     & \textbf{0.816}            & 0.513                                 & 0.138                                 & \textbf{0.015}                        \\
                                         &                   & + Polar         & \textbf{0.153}         & \textbf{0.219}         & \textbf{0.655}            & \textbf{0.384}            & \textbf{0.979}            & 0.846                     & \textbf{0.517}                        & \textbf{0.153}                        & 0.002                                 \\
                                         &                   & Rectification   & 0.169                  & 0.232                  & 0.673                     & 0.388                     & 1.004                     & 0.760                     & 0.538                                 & 0.189                                 & 0.024                                 \\
            \midrule                     &                   & Zero-Shot       & 0.034                  & 0.083                  & 0.998                     & 0.469                     & 1.031                     & 1.770                     & 0.195                                 & 0.009                                 & 0.000                                 \\
                                         &                   & Baseline        & 0.167                  & 0.216                  & 0.716                     & 0.378                     & 0.847                     & 1.216                     & 0.468                                 & \textbf{0.131}                        & \textbf{0.103}                        \\
            BEVFormer                    & ResNet-101        & DA-VTM          & 0.167                  & \textbf{0.230}         & 0.731                     & 0.361                     & \textbf{0.736}            & \textbf{1.027}            & 0.467                                 & 0.113                                 & 0.081                                 \\
                                         &                   & + Polar         & \textbf{0.175}         & 0.221                  & \textbf{0.715}            & \textbf{0.357}            & 0.858                     & 1.032                     & \textbf{0.494}                        & 0.120                                 & 0.066                                 \\
                                         &                   & Rectification   & 0.218                  & 0.275                  & 0.660                     & 0.357                     & 0.792                     & 0.862                     & 0.566                                 & 0.187                                 & 0.098                                 \\
            \midrule                     &                   & Zero-Shot       & 0.001                  & 0.039                  & 1.119                     & 0.698                     & 1.157                     & 1.125                     & 0.010                                 & 0.000                                 & 0.000                                 \\
                                         &                   & Baseline        & 0.266                  & 0.280                  & 0.668                     & 0.340                     & 0.885                     & \textbf{0.932}            & 0.597                                 & 0.260                                 & 0.046                                 \\
            PETR                         & VovNet-99         & DA-VTM          & 0.269                  & 0.283                  & \textbf{0.596}            & \textbf{0.334}            & 0.884                     & 1.057                     & \textbf{0.606}                        & 0.230                                 & \textbf{0.137}                        \\
                                         &                   & + Polar         & \textbf{0.280}         & \textbf{0.288}         & 0.598                     & 0.347                     & \textbf{0.875}            & 0.993                     & 0.602                                 & \textbf{0.296}                        & 0.104                                 \\
                                         &                   & Rectification   & 0.295                  & 0.337                  & 0.629                     & 0.337                     & 0.839                     & 0.677                     & 0.610                                 & 0.320                                 & 0.146                                 \\
            \bottomrule[2pt]
        \end{tabular}
    \end{table*}

    \paragraph{Implementation Details}
    All models are implemented within the MMDetection3D~\cite{mmdet3d2020} framework.
    We maintain the original backbone configurations for each model: VovNet-99 for
    PETR, ResNet-101 for BEVFormer (small, static version), and ResNet-50 for
    BEVDet, all pretrained on ImageNet~\cite{deng2009imagenet}. The fisheye images
    are cropped to 1408×376 resolution to maintain a unified image tensor size.
    For BEVDet, we preserve the original input height and downscale images to 960×256
    for a fair comparison with baselines. All models are trained in a distributed manner for
    24 epochs with a batch size of 8, using the AdamW optimizer~\cite{loshchilov2017decoupled}
    with an initial learning rate scaled according to the number of used NVIDIA A5000
    GPUs, decayed using a cosine annealing schedule. Beyond the core architectural
    changes described in Section~\ref{sec:methodology}, all other
    hyperparameters follow the default configurations for the respective models
    in the MMDetection3D library to isolate the impact of the introduced
    adaptations. Evaluation is performed on 10 classes: [car, truck, trailer,
    bus, bicycle, motorcycle, pedestrian, pole, object, traffic sign] using nuScenes
    benchmark metrics without average attribute error (AAE) and rebalanced weights
    for NDS calculation.

    \subsection{Main Results and Analysis}
    \paragraph{Zero-Shot Performance and Rectification}
    The zero-shot results in Table~\ref{tab:comprehensive_results} reveal the
    severe impact of geometric and sensor setup mismatch when applying nuScenes-trained
    models directly to fisheye data. All models exhibit catastrophic performance
    degradation, with PETR and BEVDet achieving near-zero mAP and BEVFormer
    slightly better at 0.034 mAP. This confirms that models trained exclusively on
    pinhole camera images suffer from severe data domain gap.

    The rectification results establish performance upper bounds. PETR achieves
    the highest overall performance (0.295 mAP, 0.337 NDS), followed by
    BEVFormer and BEVDet. Per-class AP reveals strong car detection (0.538–0.610)
    but poor results for underrepresented classes like buses (\textless0.1) and
    small-scale objects like pedestrians (0.2–0.3), especially for projection-based
    methods. PETR shows a more balanced performance, suggesting projection-free architectures
    handle class imbalance better. The dataset’s extreme skew toward static
    infrastructure (buildings: 572K, poles: 375K) versus dynamic objects (cars: 430K,
    buses: 1K) biases projection-based models toward dominant classes, reducing overall
    performance.

    \paragraph{Distortion-Aware View Transformation}
    When trained directly on fisheye data (Baseline) while ignoring distortion
    modeling, performance varies significantly across architectures. PETR
    achieves the highest baseline performance (0.266 mAP, 0.280 NDS), followed
    by BEVFormer (0.167 mAP, 0.216 NDS), while BEVDet shows more modest results
    (0.121 mAP, 0.159 NDS) with slightly lower input resolution. This ranking
    suggests that projection-free architectures (PETR) are inherently more
    robust to geometric inconsistency than other methods, as they can partially
    compensate for distortion through learned representations rather than explicit
    geometric transformations. These results confirm that the pinhole assumption
    embedded in these architectures fundamentally breaks down when confronted
    with fisheye distortion, leading to incorrect feature projection and poor spatial
    reasoning.

    Integrating the MEI camera model (DA-VTM) produces varied results across architectures.
    PETR shows modest improvement (0.269 mAP, 0.283 NDS), benefiting from
    geometrically accurate 3D positional encoding. BEVFormer maintains similar mAP
    (0.167) but achieves better NDS (0.230) through improved True Positive metrics.
    Most notably, BEVDet demonstrates significant gains (0.150 mAP, 0.212 NDS), a
    24\% relative mAP improvement over baseline, suggesting that explicit distortion
    modeling particularly benefits depth-based approaches. Examining error
    metrics, models show reduced translation error (mATE) with DA-VTM,
    confirming better spatial localization after proper distortion handling.

    \paragraph{Polar Coordinate Enhancement}
    Adding polar coordinate representations (+ Polar) consistently improves all models,
    with PETR achieving the best overall performance (0.280 mAP, 0.288 NDS),
    while BEVFormer (0.175 mAP, 0.221 NDS) and BEVDet (0.153 mAP, 0.219 NDS)
    also show clear benefits. Per-class analysis reveals that polar coordinates particularly
    enhance pedestrian detection, with PETR achieving 0.296 AP (vs. 0.230) and BEVDet
    reaching 0.153 AP (vs. 0.138). This confirms our hypothesis that polar coordinates
    naturally align with fisheye geometry, providing more intuitive spatial
    representations for transformer-based reasoning about radially distorted
    images, especially for smaller objects that suffer more from distortion effects.

    \subsection{Robustness Analysis}
    \begin{figure}[htbp]
        \centering
        \includegraphics[width=1.0\linewidth]{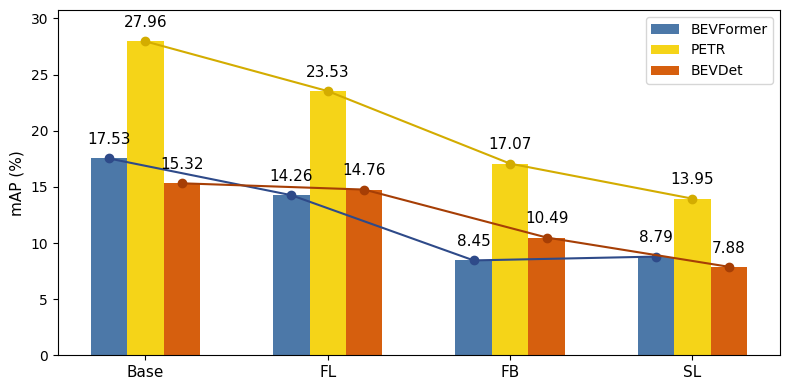}
        \caption{Model robustness under camera failure scenarios. FL, FB and SL
        denote the front-left, front-both and side-left.}
        \label{fig:camera_robustness}
    \end{figure}
    \paragraph{Robustness under camera failure}
    Figure~\ref{fig:camera_robustness} shows the mAP performance under different
    camera loss scenarios. PETR consistently outperforms other models across all
    configurations, with BEVFormer and BEVDet showing different robustness
    patterns. When losing the front-left camera (FL), BEVDet shows remarkable
    resilience with only a 3.7\% drop, while BEVFormer and PETR experience
    larger decreases. With both front cameras disabled (FB), PETR maintains 61.1\%
    of its baseline performance, demonstrating stronger fisheye-only perception
    capabilities than others. Side-left camera failure (SL) significantly impacts
    all models with similar severity, causing approximately 50\% performance drops
    across architectures. This suggests side cameras provide critical
    information that cannot be fully compensated by remaining views. PETR's consistent
    superiority confirms that projection-free architectures better handle geometric
    uncertainty in mixed camera configurations.

    \begin{table}[t]
        \centering
        \caption{mAP (\%) across distance ranges on KITTI-360. (rect) for
        models on rectified images and (ours) represent distortion-aware models on mixed
        inputs.}
        \label{tab:range_map} \resizebox{\linewidth}{!}{
        \begin{tabular}{lccccc}
            \toprule Method          & 0--10\,m & 10--20\,m & 20--30\,m & 30--40\,m & 40--50\,m \\
            \midrule BEVFormer(rect) & 49.54    & 26.08     & 7.87      & 2.17      & 0.63      \\
            BEVFormer (ours)         & 47.38    & 20.17     & 5.43      & 1.83      & 0.40      \\
            \midrule PETR(rect)      & 55.87    & 37.70     & 14.95     & 4.83      & 1.45      \\
            PETR (ours)              & 56.57    & 35.52     & 12.49     & 4.50      & 1.44      \\
            \midrule BEVDet(rect)    & 35.17    & 19.90     & 6.41      & 2.35      & 0.57      \\
            BEVDet (ours)            & 34.22    & 18.90     & 5.64      & 1.65      & 0.36      \\
            \bottomrule
        \end{tabular}
        }
    \end{table}
    \paragraph{Performance along distance ranges}
    Table~\ref{tab:range_map} reports mAP across distance bins for three BEV
    architectures. For all models, accuracy declines rapidly with range, highlighting
    the challenge of long-range perception in fisheye views. Training with
    rectified images excels at nearly all distances over distortion-aware models, as
    rectification restores pinhole-like geometry that these architectures were originally
    designed for. However, distortion-aware PETR achieves the best near-field score
    (56.57\% 0--10\,m), slightly higher than on rectified images, while both BEVFormer
    and BEVDet show slightly lower performance with distortion modeling than with
    rectification (-2.16\% and -0.95\% respectively). Beyond 10m, rectified models
    consistently outperform their distortion-aware counterparts across all architectures,
    with the performance gap widening at longer ranges. These results suggest that
    distortion-aware modeling preserves near-field detail effectively but struggles
    with long-range depth estimation and spatial reasoning.

    \paragraph{Angular-stratified performance}
    Table~\ref{tab:angular_map} compares baseline and distortion-aware models
    across three azimuthal sectors: front ($120^{\circ}$, primarily covered by pinhole
    cameras), back ($120^{\circ}$, severe fisheye distortion), and sides ($120^{\circ}$,
    moderate distortion). Distortion-aware adaptations yield clear improvements in
    regions with higher distortion, like side and back areas. 
    Interestingly, only BEVFormer shows a slight decrease in front-facing mAP
    (-0.06), while all other models demonstrate improvements across all sectors, confirming that unified MEI camera modeling does not degrade performance on
    pinhole views.
    BEVDet demonstrates the most substantial relative improvements,
    with gains of +3.15 mAP (+21.8\%) in front, +4.24 mAP (+54.3\%) in back, and
    +3.63 mAP (+26.6\%) in side regions. PETR, despite starting from a
    higher baseline, still achieves meaningful gains, particularly in front
    regions (+1.51). The consistent improvements in the side and back regions across all architectures
    confirm that proper distortion modeling is particularly critical for areas
    where fisheye distortion is most pronounced, enabling more accurate spatial
    reasoning for 3D object detection.

    \begin{table}[t]
        \centering
        \caption{Angular-stratified mAP (\%) on KITTI-360. Baseline for standard
        training on mixed images and (ours) = distortion-aware models.}
        \label{tab:angular_map} \resizebox{\linewidth}{!}{
        \begin{tabular}{lccc}
            \toprule Method             & Front $120^{\circ}$ & Back $120^{\circ}$ & Sides $120^{\circ}$ \\
            \midrule BEVFormer baseline & \textbf{19.42}      & 12.56              & 18.08               \\
            BEVFormer (ours)            & 19.36               & \textbf{14.06}     & \textbf{21.44}      \\
            \midrule PETR baseline      & 29.84               & 21.87              & 28.77               \\
            PETR (ours)                 & \textbf{31.35}      & \textbf{22.68}     & \textbf{29.84}      \\
            \midrule BEVDet baseline    & 14.43               & 7.81              & 13.67               \\
            BEVDet (ours)               & \textbf{17.58}      & \textbf{12.05}     & \textbf{17.30}      \\
            \bottomrule
        \end{tabular}
        }
    \end{table}

    \section{CONCLUSIONS}
    \label{sec:conclusion}

    This work addresses the critical gap between mixed pinhole-fisheye camera configurations
    and BEV 3D object detection with distortion-aware adaptations and enhancements.
    Through systematic evaluation on our converted KITTI-360 benchmark for multi-view
    3DOD, we demonstrate that directly applying pinhole-trained models to an out-of-distribution
    dataset results in catastrophic failure with near-zero mAP. Our analysis
    reveals architectural differences in handling fisheye distortion. Projection-free
    methods (PETR) prove most adaptable, achieving highest mAP with distortion-aware
    3D positional encoding along the polar coordinate enhancement. Backward-projection
    approaches (BEVFormer) show moderate gains. While starting from a lower baseline, forward-projection methods (BEVDet) show the most 
    dramatic relative gains from our adaptations. These trends suggest that decoupling feature learning
    from explicit geometry improves adaptability to sensor variation.
    Introducing polar positional embeddings consistently benefits all methods, confirming
    that aligning BEV grids with the radial nature of fisheye distortion is a powerful
    geometric prior. Our robustness analysis confirms that PETR maintains
    superior performance even under camera failure scenarios, retaining viability
    with fisheye-only perception. These findings establish architectural
    guidelines for fisheye-based 3D detection: projection-free designs are preferable
    to depth-dependent approaches, and geometric priors should align with sensor
    characteristics. While our work demonstrates the feasibility of distortion-aware
    BEV detection, significant challenges remain, particularly for small object detection in highly
    distorted image regions and handling severely imbalanced sample distributions
    that bias models toward dominant classes.

    \addtolength{\textheight}{-12cm} 








    \bibliographystyle{ieeetr}
    \bibliography{IEEEfull}

@article{kumar2023surround,
  title={Surround-view fisheye camera perception for automated driving: Overview, survey \& challenges},
  author={Kumar, Varun Ravi and Eising, Ciar{\'a}n and Witt, Christian and Yogamani, Senthil Kumar},
  journal={IEEE Transactions on Intelligent Transportation Systems},
  volume={24},
  number={4},
  pages={3638--3659},
  year={2023},
  publisher={IEEE}
}

@inproceedings{caesar2020nuscenes,
  title={nuscenes: A multimodal dataset for autonomous driving},
  author={Caesar, Holger and Bankiti, Varun and Lang, Alex H and Vora, Sourabh and Liong, Venice Erin and Xu, Qiang and Krishnan, Anush and Pan, Yu and Baldan, Giancarlo and Beijbom, Oscar},
  booktitle={Proceedings of the IEEE/CVF conference on computer vision and pattern recognition},
  pages={11621--11631},
  year={2020}
}

@inproceedings{samani2023f2bev,
  title={F2BEV: Bird's Eye View Generation from Surround-View Fisheye Camera Images for Automated Driving},
  author={Samani, Ekta U and Tao, Feng and Dasari, Harshavardhan R and Ding, Sihao and Banerjee, Ashis G},
  booktitle={2023 IEEE/RSJ International Conference on Intelligent Robots and Systems (IROS)},
  pages={9367--9374},
  year={2023},
  organization={IEEE}
}

@article{liao2022kitti,
  title={Kitti-360: A novel dataset and benchmarks for urban scene understanding in 2d and 3d},
  author={Liao, Yiyi and Xie, Jun and Geiger, Andreas},
  journal={IEEE Transactions on Pattern Analysis and Machine Intelligence},
  volume={45},
  number={3},
  pages={3292--3310},
  year={2022},
  publisher={IEEE}
}

@article{li2022bevsurvey,
  author={Li, Hongyang and Sima, Chonghao and Dai, Jifeng and Wang, Wenhai and Lu, Lewei and Wang, Huijie and Zeng, Jia and Li, Zhiqi and Yang, Jiazhi and Deng, Hanming and Tian, Hao and Xie, Enze and Xie, Jiangwei and Chen, Li and Li, Tianyu and Li, Yang and Gao, Yulu and Jia, Xiaosong and Liu, Si and Shi, Jianping and Lin, Dahua and Qiao, Yu},
  journal={IEEE Transactions on Pattern Analysis and Machine Intelligence}, 
  title={Delving Into the Devils of Bird's-Eye-View Perception: A Review, Evaluation and Recipe}, 
  year={2023},
  volume={},
  number={},
  pages={1-20},
  doi={10.1109/TPAMI.2023.3333838}
}

@inproceedings{liu2022petr,
  title={Petr: Position embedding transformation for multi-view 3d object detection},
  author={Liu, Yingfei and Wang, Tiancai and Zhang, Xiangyu and Sun, Jian},
  booktitle={European conference on computer vision},
  pages={531--548},
  year={2022},
  organization={Springer}
}

@article{li2024bevformer,
  title={Bevformer: learning bird's-eye-view representation from lidar-camera via spatiotemporal transformers},
  author={Li, Zhiqi and Wang, Wenhai and Li, Hongyang and Xie, Enze and Sima, Chonghao and Lu, Tong and Yu, Qiao and Dai, Jifeng},
  journal={IEEE Transactions on Pattern Analysis and Machine Intelligence},
  year={2024},
  publisher={IEEE}
}

@article{huang2021bevdet,
  title={Bevdet: High-performance multi-camera 3d object detection in bird-eye-view},
  author={Huang, Junjie and Huang, Guan and Zhu, Zheng and Ye, Yun and Du, Dalong},
  journal={arXiv preprint arXiv:2112.11790},
  year={2021}
}

@inproceedings{mei2007single,
  title={Single view point omnidirectional camera calibration from planar grids},
  author={Mei, Christopher and Rives, Patrick},
  booktitle={Proceedings 2007 IEEE International Conference on Robotics and Automation},
  pages={3945--3950},
  year={2007},
  organization={IEEE}
}

@inproceedings{jiang2023polarformer,
  title={Polarformer: Multi-camera 3d object detection with polar transformer},
  author={Jiang, Yanqin and Zhang, Li and Miao, Zhenwei and Zhu, Xiatian and Gao, Jin and Hu, Weiming and Jiang, Yu-Gang},
  booktitle={Proceedings of the AAAI conference on Artificial Intelligence},
  volume={37},
  number={1},
  pages={1042--1050},
  year={2023}
}

@article{yu2024polarbevdet,
  title={PolarBEVDet: Exploring Polar Representation for Multi-View 3D Object Detection in Bird's-Eye-View},
  author={Yu, Zichen and Liu, Quanli and Wang, Wei and Zhang, Liyong and Zhao, Xiaoguang},
  journal={arXiv preprint arXiv:2408.16200},
  year={2024}
}

@article{wang2023focal,
  title={Focal-petr: Embracing foreground for efficient multi-camera 3d object detection},
  author={Wang, Shihao and Jiang, Xiaohui and Li, Ying},
  journal={IEEE Transactions on Intelligent Vehicles},
  volume={9},
  number={1},
  pages={1481--1489},
  year={2023},
  publisher={IEEE}
}

@article{huang2022bevdet4d,
  title={Bevdet4d: Exploit temporal cues in multi-camera 3d object detection},
  author={Huang, Junjie and Huang, Guan},
  journal={arXiv preprint arXiv:2203.17054},
  year={2022}
}

@inproceedings{wang2023exploring,
  title={Exploring object-centric temporal modeling for efficient multi-view 3d object detection},
  author={Wang, Shihao and Liu, Yingfei and Wang, Tiancai and Li, Ying and Zhang, Xiangyu},
  booktitle={Proceedings of the IEEE/CVF international conference on computer vision},
  pages={3621--3631},
  year={2023}
}

@inproceedings{yang2023bevformer,
  title={Bevformer v2: Adapting modern image backbones to bird's-eye-view recognition via perspective supervision},
  author={Yang, Chenyu and Chen, Yuntao and Tian, Hao and Tao, Chenxin and Zhu, Xizhou and Zhang, Zhaoxiang and Huang, Gao and Li, Hongyang and Qiao, Yu and Lu, Lewei and others},
  booktitle={Proceedings of the IEEE/CVF conference on computer vision and pattern recognition},
  pages={17830--17839},
  year={2023}
}

@inproceedings{li2023fb,
  title={Fb-bev: Bev representation from forward-backward view transformations},
  author={Li, Zhiqi and Yu, Zhiding and Wang, Wenhai and Anandkumar, Anima and Lu, Tong and Alvarez, Jose M},
  booktitle={Proceedings of the IEEE/CVF International Conference on Computer Vision},
  pages={6919--6928},
  year={2023}
}

@inproceedings{philion2020lift,
  title={Lift, splat, shoot: Encoding images from arbitrary camera rigs by implicitly unprojecting to 3d},
  author={Philion, Jonah and Fidler, Sanja},
  booktitle={European conference on computer vision},
  pages={194--210},
  year={2020},
  organization={Springer}
}

@article{kannala2006generic,
  title={A generic camera model and calibration method for conventional, wide-angle, and fish-eye lenses},
  author={Kannala, Juho and Brandt, Sami S},
  journal={IEEE transactions on pattern analysis and machine intelligence},
  volume={28},
  number={8},
  pages={1335--1340},
  year={2006},
  publisher={IEEE}
}

@inproceedings{carlsson2024heal,
  title={Heal-swin: A vision transformer on the sphere},
  author={Carlsson, Oscar and Gerken, Jan E and Linander, Hampus and Spie{\ss}, Heiner and Ohlsson, Fredrik and Petersson, Christoffer and Persson, Daniel},
  booktitle={Proceedings of the IEEE/CVF Conference on Computer Vision and Pattern Recognition},
  pages={6067--6077},
  year={2024}
}

@inproceedings{athwale2023darswin,
  title={Darswin: Distortion aware radial swin transformer},
  author={Athwale, Akshaya and Afrasiyabi, Arman and Lag{\"u}e, Justin and Shili, Ichrak and Ahmad, Ola and Lalonde, Jean-Fran{\c{c}}ois},
  booktitle={Proceedings of the IEEE/CVF International Conference on Computer Vision},
  pages={5929--5938},
  year={2023}
}

@article{griffiths2024adapting,
  title={Adapting CNNs for fisheye cameras without retraining},
  author={Griffiths, Ryan and Dansereau, Donald G},
  journal={arXiv preprint arXiv:2404.08187},
  year={2024}
}

@article{liu2022bevfusion,
  title={Bevfusion: Multi-task multi-sensor fusion with unified bird's-eye view representation},
  author={Liu, Zhijian and Tang, Haotian and Amini, Alexander and Yang, Xinyu and Mao, Huizi and Rus, Daniela and Han, Song},
  journal={arXiv preprint arXiv:2205.13542},
  year={2022}
}

@inproceedings{yogamani2019woodscape,
  title={Woodscape: A multi-task, multi-camera fisheye dataset for autonomous driving},
  author={Yogamani, Senthil and Hughes, Ciar{\'a}n and Horgan, Jonathan and Sistu, Ganesh and Varley, Padraig and O'Dea, Derek and Uric{\'a}r, Michal and Milz, Stefan and Simon, Martin and Amende, Karl and others},
  booktitle={Proceedings of the IEEE/CVF International Conference on Computer Vision},
  pages={9308--9318},
  year={2019}
}

@article{sekkat2022synwoodscape,
  title={SynWoodScape: Synthetic surround-view fisheye camera dataset for autonomous driving},
  author={Sekkat, Ahmed Rida and Dupuis, Yohan and Kumar, Varun Ravi and Rashed, Hazem and Yogamani, Senthil and Vasseur, Pascal and Honeine, Paul},
  journal={IEEE Robotics and Automation Letters},
  volume={7},
  number={3},
  pages={8502--8509},
  year={2022},
  publisher={IEEE}
}

@inproceedings{sun2020scalability,
  title={Scalability in perception for autonomous driving: Waymo open dataset},
  author={Sun, Pei and Kretzschmar, Henrik and Dotiwalla, Xerxes and Chouard, Aurelien and Patnaik, Vijaysai and Tsui, Paul and Guo, James and Zhou, Yin and Chai, Yuning and Caine, Benjamin and others},
  booktitle={Proceedings of the IEEE/CVF conference on computer vision and pattern recognition},
  pages={2446--2454},
  year={2020}
}

@inproceedings{liu2023petrv2,
  title={Petrv2: A unified framework for 3d perception from multi-camera images},
  author={Liu, Yingfei and Yan, Junjie and Jia, Fan and Li, Shuailin and Gao, Aqi and Wang, Tiancai and Zhang, Xiangyu},
  booktitle={Proceedings of the IEEE/CVF international conference on computer vision},
  pages={3262--3272},
  year={2023}
}

@misc{mmdet3d2020,
    title={{MMDetection3D: OpenMMLab} next-generation platform for general {3D} object detection},
    author={MMDetection3D Contributors},
    howpublished = {\url{https://github.com/open-mmlab/mmdetection3d}},
    year={2020}
}

@inproceedings{deng2009imagenet,
  title={Imagenet: A large-scale hierarchical image database},
  author={Deng, Jia and Dong, Wei and Socher, Richard and Li, Li-Jia and Li, Kai and Fei-Fei, Li},
  booktitle={2009 IEEE conference on computer vision and pattern recognition},
  pages={248--255},
  year={2009},
  organization={Ieee}
}

@article{loshchilov2017decoupled,
  title={Decoupled weight decay regularization},
  author={Loshchilov, Ilya and Hutter, Frank},
  journal={arXiv preprint arXiv:1711.05101},
  year={2017}
}
\end{document}